\documentclass[11pt,a4paper]{article}
\usepackage{times}
\usepackage{latexsym}
\usepackage{amssymb}
\usepackage{amsmath}
\usepackage{url}
\usepackage{breqn}
\usepackage{graphicx}
\usepackage{graphics}

\title{Generalized Attention Mechanism and Relative Position for Transformer}
\author{R. V. R. Pandya \\ rajavikram.pandya@outlook.com \\ July 22, 2022} 
\date{}
\begin{document}

\maketitle

\begin{abstract}
In this paper, we propose generalized attention mechanism (GAM) by first suggesting a new interpretation for self-attention mechanism of Vaswani et al. \cite{vaswani2017attention}. Following the interpretation, we provide description for different variants of attention mechanism which together form GAM. Further, we propose a new relative position representation within the framework of GAM. This representation can be easily utilized for cases in which elements next to each other in input sequence can be at random locations in actual dataset/corpus.   
\end{abstract}

\section{Introduction}
Vaswani et al. \cite{vaswani2017attention} proposed self-attention  mechanism based neural network for sequence transduction, namely Transformer, as computationally efficient alternative to recurrent and convolutional neural networks. Self-attention mechanism has been interpreted in terms of query, key and value of different elements of the sequence. Their work includes absolute position representation of these elements through sine and cosine functions. Later Shaw et al. \cite{shaw2018self} improved it by including relative position representation for different elements of sequence. Since then, within the framework of self-attention mechanism, different models have been suggested for relative position representation (see \cite{dufter2021position,ke2020rethinking} and references cited therein).  

In this paper, we first describe self-attention mechanism as suggested by Vaswani et al. \cite{vaswani2017attention} using tensor notations. We use Einstein summation convention (i.e. summation over repeated indices) while writing various equations in tensor notations. An alternate interpretation for attention mechanism is suggested which does not require query and key terminology. Then we provide details of our generalized attention mechanism and inclusion of relative position.

\section{Self-Attention Mechanism of Vaswani et al. \cite{vaswani2017attention, shaw2018self}}

Consider input sequence ${\bf Y}$ of $n$ elements ${\bf y}^\alpha$ where superscript $\alpha = 1,2,3,\ldots, n$ represents different elements of the sequence. For language processing each ${\bf y}^\alpha$  is a vector corresponding to a word/token ${ \nu}^\alpha$. ${\bf Y}$ can be written as
\begin{eqnarray}
	{\bf Y} &=&({\nu}^1, {\nu}^2,\dots, {\nu}^\alpha, \ldots, {\nu}^n)\nonumber \\
	&=&({\bf y}^1, {\bf y}^2,\dots, {\bf y}^\alpha, \ldots, {\bf y}^n). \label{eq:vY}
\end{eqnarray}

We use Einstein summation convention (i.e. summation over repeated indices) while writing various equations in tensor notations in this paper. Each element ${\bf y}^\alpha$ is tensor of rank $R \ge 0$ and in present case of language processing we limit our discussion to element being $M$-dimensional vector (tensor of $R = 1$), written as
\begin{equation}\label{eq:vyalpha}
	{\bf y}^\alpha = (y^\alpha_1, y^\alpha_2, \ldots, y^\alpha_M ) \equiv y^\alpha_i\, \mbox{where} \,\, \{i = 1, 2, \ldots, M\}.
\end{equation}

Similarly, the output sequence ${\bf Z}$ of $n$ elements ${\bf z}^\alpha$ ($\alpha = {1,2,3, \ldots,n}$) can be written as
\begin{equation}\label{eq:vzalpha}
	{\bf z}^\alpha = (z^\alpha_1, z^\alpha_2, \ldots, z^\alpha_{M_v} ) \equiv z^\alpha_i\, \mbox{where} \,\, \{i = 1, 2, \ldots, M_v\}.
\end{equation}
For multi-head case, later we use ${^a}z^\alpha_i$ where superscript $a = {1, 2, \ldots, h} $ on $z$ is used to represent output ${\bf z}^\alpha$ in particular head $a$ from self-attention sub-layer. 

The three parameter matrices $^a{\bf W}^Q$, $^a{\bf W}^K$, $^a{\bf W}^V$ for query ($Q$), key ($K$) and value ($V$), respectively, for a particular head $a$ are written using tensor notation as
\begin{eqnarray}
	^a{\bf W}^Q = {^a}{W}^Q_{ij},\label{vpm1}\\
	^a{\bf W}^K = {^a}W^K_{ij},\label{vpm2}\\
	^a{\bf W}^V = {^a}W^V_{ij}, \label{vpm3}\\
\end{eqnarray}
where superscript $a = {1, 2, \ldots, h} $ on $W$ is used to represent any particular head $a$ in $h$ multi-head. Also, subscripts $i$ and $j$ are used to represent different elements of matrices.

The $j$th component of output vector from self-attention sub-layer of head $a$, corresponding to element $\alpha$ of input sequence, is given by  
\begin{equation}\label{eq:vzout}
	{^a}z^\alpha_j = \sum_{\beta =1}^n \,{^a}\Phi^{\alpha \beta}\,y^\beta_i\,({^a}W^V_{ij}).
\end{equation}
The weight coefficient $^a\Phi^{\alpha \beta}$ is given by
\begin{equation}\label{eq:valphaij}
	^a\Phi^{\alpha \beta}= \frac{\exp (^ae^{\alpha \beta})}{\sum_{\gamma = 1}^n \exp (^ae^{\alpha \gamma})}.
\end{equation}
Also  
\begin{equation}\label{eq:veij}
	^ae^{\alpha \beta}=\frac{1}{\sqrt d_k}\Bigl\{y^\alpha_r\,({^a}{W}^Q_{rs})\Bigr\}\,\Bigl\{y^\beta_t\,({^a}{W}^K_{tu})\Bigr\} \delta_{su},
\end{equation}
where $\delta_{su}$ is Kronecker delta ($\delta_{su} = 1$ when $s=u$ and $\delta_{su} = 0$ when $s\ne u$). It should be noted that summation is implied on repeated indices in Eq. (\ref{eq:veij}).
  
\section{Present Generalized Attention Mechanism (GAM)} 
Consider Eq. (\ref{eq:veij}) which can be also written as
\begin{equation}\label{eq:myveij}
	^ae^{\alpha \beta}=\Bigl\{y^\alpha_r\,y^\beta_t\Bigr\}\,{^a}{B}_{rt},
\end{equation} 
where
\begin{equation}\label{eq:myveijB}
	{^a}B_{rt}=\Bigl\{\frac{1}{\sqrt d_k} ({^a}{W}^Q_{rs})\,({^a}{W}^K_{tu})\,\delta_{su}\Bigr\}.
\end{equation} 
It should be noted in view of Eq. (\ref{eq:myveij}) that self-attention mechanism of Vaswani et al. can be interpreted as combination of higher order features of input (i.e. $y^\alpha_r\,y^\beta_t$) and square matrix ${^a}{\bf B}$ whose components are denoted by ${^a}B_{rt}$. These higher order features along with ${^a}{\bf B}$ are responsible for evolving value vectors of elements of input sequence.   The dot-product attention of Vaswani et al. can be completely described (for head $a$) using two parameter matrices ${^a}{\bf B}$, $^a{\bf W}^V$ instead of three parameter matrices $^a{\bf W}^Q$, $^a{\bf W}^K$, $^a{\bf W}^V$. And interpretation of their attention mechanism in terms of query and key may be abandoned.    

We base generalized attention model (GAM) on Eqs. (\ref{eq:vzout},\ref{eq:valphaij},\ref{eq:myveij}) requiring learnable parameter matrices ${^a}{\bf B}$ and $^a{\bf W}^V$. The GAM equations can be written as 

\begin{equation}\label{eq:GAM1}
	{^a}z^\alpha_j = \sum_{\beta =1}^n \,{^a}\Phi^{\alpha \beta}\,y^\beta_i\,({^a}W^V_{ij}),
\end{equation}

\begin{equation}\label{eq:GM2}
	^a\Phi^{\alpha \beta}= \sum_{i=1}^{N_B}W^{P}_i\,({^a}\Psi^{\alpha \beta}_i),
\end{equation}
where 
\begin{equation}\label{eq:GM3}
	{^a}\Psi^{\alpha \beta}_i = \frac{\exp ({^a}\epsilon ^{\alpha \beta}_i)}{\sum_{\gamma = 1}^n \exp ({^a}\epsilon^{\alpha \gamma}_i)},
\end{equation}

\begin{equation}\label{eq:GM4} 
{^a}\epsilon ^{\alpha \beta}_i = f(y^\alpha_r\,y^\beta_t)\,\Bigl[{^{(a,i)}}{B}_{rt}\Bigr],
\end{equation}
and $N_B$ are number of different parameter matrix ${^{(a,i)}}{\bf B}$ in the same attention head. Each ${^{(a,i)}}{\bf B}$ of head $a$ can be think of as different portion of 'brain' in the head.  Also $f(\ldots)$ represents function of $y^\alpha_r\,y^\beta_t$ and superscript $(a,i)$ on the left of $B$ represents $i$th parameter matrix ${^{(a,i)}}{\bf B}$  in attention head $a$. Two functional form for  $f(y^\alpha_r\,y^\beta_t)$ of power law and polynomial types can be considered for GAM and which are written as 

Power law type:
\begin{equation}\label{eq:GM5} 
		f(y^\alpha_r\,y^\beta_t) = (y^\alpha_ry^\beta_t)^{n_1}, \,\, n_1 > 0,
\end{equation}

Polynomial type:
\begin{equation}\label{eq:GM52} 
	f(y^\alpha_r\,y^\beta_t) = \sum_{l = 1}^L A_l (y^\alpha_ry^\beta_t)^l.
\end{equation}
where $A_l$'s are learnable  scalar parameters. 

Also, $W^P_i$ can be either considered equal to $1/N_B$ or following constraint can be utilized during learning:
\begin{equation}
\sum_{i=1}^{N_B} W^P_i =1.
\end{equation}
\section{Relative Position in GAM} 
Different models exist for inclusion of relative position representation within the framework of self-attention mechanism, for example see references \cite{shaw2018self,raffel2020exploring,dufter2021position,ke2020rethinking}. Here we suggest yet another model to include effect of relative position of input elements within the framework of GAM.

In this section, we assume that position related information are not included in ${\bf y}^\alpha$ of GAM equations which are written above. Linear combination of contributions from relative position ${^a}\pi^\alpha_j$ and ${^a}z^\alpha_j$ become output of GAM and can be written as
\begin{equation}\label{eq:ztotal}
[{^a}z^\alpha_j]_{total} =  (c_1) \,({^a}z^\alpha_j) +  (1-c_1)\,({^a}\pi^\alpha_j), \,\, 0<c_1<1,
\end{equation}
where 
\begin{equation}\label{eq:piout}
	{^a}\pi^\alpha_j = \sum_{\beta =1}^n \,{^a}\Theta^{\alpha \beta}\,y^\beta_i\,({^a}W^V_{ij}),
\end{equation}
\begin{equation}\label{eq:GM2p}
	^a\Theta^{\alpha \beta}= \sum_{i=1}^{N_B}W^{S}_i\,({^a}\xi^{\alpha \beta}_i),
\end{equation}
where 
\begin{equation}\label{eq:GM3p}
	{^a}\xi^{\alpha \beta}_i = \frac{\exp ({^a}\delta ^{\alpha \beta}_i)}{\sum_{\gamma = 1}^n \exp ({^a}\delta^{\alpha \gamma}_i)}
\end{equation}
\begin{equation}\label{eq:GM4p} 
	{^a}\delta ^{\alpha \beta}_i = f(p^\alpha_r\,p^\beta_t)\,\Bigl[{^{(a,i)}}{B}^P_{rt}\Bigr].
\end{equation}
The function $f(\ldots)$ of $p^\alpha_r\,p^\beta_t$ can be considered as power law type (Eq. \ref{eq:GM5}) or polynomial type (Eq. \ref{eq:GM52}).
Here constraint on 
Another possibility for $[{^a}z^\alpha_j]_{total}$ which can be explored is geometric average, written as
\begin{equation}\label{eq:ztotal2}
	[{^a}z^\alpha_j]_{total} =  \sqrt{({^a}z^\alpha_j)({^a}\pi^\alpha_j)}.
\end{equation}

Now we discuss methodology to obtain relative position vector $p^\alpha_r$.
Consider ${\bf p}^\alpha$ as embedded relative position vector corresponding to input element ${\bf y}^\alpha$.  The dimension of ${\bf p}^\alpha$ is identical to that of ${\bf y}^\alpha$ and is equal to $M$. The embedded vector can be written as
\begin{equation}\label{eq:vpalpha}
	{\bf p}^\alpha = (p^\alpha_1, p^\alpha_2, \ldots, p^\alpha_M ) \equiv p^\alpha_i\, \mbox{where} \,\, \{i = 1, 2, \ldots, M\}.
\end{equation}
The embedded vector for different $\alpha = 1,2,\ldots, n$ can be learned during training from known input relative position vector ${\bf r}^\alpha$ whose dimension is equal to $n$. ${\bf r}^\alpha$ can be written as
\begin{equation}\label{eq:vralpha}
{\bf r}^\alpha = (r^\alpha_1, r^\alpha_2, \ldots, r^\alpha_n ) \equiv r^\alpha_i\, \mbox{where} \,\, \{i = 1, 2, \ldots, n\},
\end{equation}
where
\begin{equation}\label{eq:vpalphav1}
	r^\alpha_\alpha = 1, 
\end{equation}
\begin{equation}\label{eq:vpalphav2}
r^\alpha_{i} = 2+ n_e\,\, \mbox{when}\,\, \alpha \ne i.
\end{equation}
Here $n_e$ is number of elements (in actual dataset/corpus) between elements at location $\alpha$ and $i$ of input sequence ${\bf Y}$. For example, consider actual corpus as 
\begin{equation}
\mbox{My name is Vikram}
\end{equation}
and $n=3$ for input sequence ${\bf Y}$. When
\begin{eqnarray}
	{\bf Y} &= & ({\nu}^1, {\nu}^2, {\nu}^3), \\
	& =& (My, name, is),
\end{eqnarray}
input relative position vectors ${\bf r}^\alpha$ can be written as
\begin{eqnarray}
	{\bf r}^1 =  (r^1_1, r^1_2,r^1_3) = (1,2,3), \\
	{\bf r}^2 = (r^2_1, r^2_2,r^2_3) =(2,1,2), \\
	{\bf r}^3 = (r^3_1, r^3_2,r^3_3) =(3,2,1). \\
\end{eqnarray}
And when 
\begin{eqnarray}
	{\bf Y} &= &({\bf y}^1, {\bf y}^2, {\bf y}^3) \\
	& =& (My, name, Vikram),
\end{eqnarray}
${\bf r}^\alpha$ can be written as
\begin{eqnarray}
	{\bf r}^1 = (r^1_1, r^1_2,r^1_3) =(1,2,4), \\
	{\bf r}^2 = (r^2_1, r^2_2,r^2_3) =(2,1,3), \\
	{\bf r}^3 = (r^3_1, r^3_2,r^3_3) =(4,3,1). \\
\end{eqnarray}

\section{Conclusion}
We have proposed generalized attention mechanism (GAM) which also includes a new way of representing relative position of elements in actual dataset/corpus. In doing so, we have suggested different interpretation for attention mechanism which abandons requirement of query and key. In GAM, similar to self-attention mechanism of Vaswani et al. \cite{vaswani2017attention},  initial static vector representation of various elements of input sequence are transformed into value vectors. These value vectors evolve into dynamic representations $[{^a}z^\alpha_j]_{total}$ under the influence of interactions among different elements of the sequence and their relative positions. These interactions are quantified in terms of higher order features of input elements, their relative positions and parameter matrices ${^{(a,i)}}{\bf B}$. The study on performance of GAM for various experiments of language processing and comparison with results of self-attention mechanism of Vaswani et al. \cite{vaswani2017attention,shaw2018self} will be performed in near future. Also, the application of GAM to time series analysis will be explored in detail.    
\bibliographystyle{plain}

\end{document}